\documentclass{article}
\usepackage{spconf,amsmath,graphicx}
\usepackage{amsfonts}
\usepackage{booktabs}
\usepackage{subcaption}
\usepackage{url,caption}
\usepackage[colorlinks]{hyperref}
\title{Unsupervised Polyglot Text-to-Speech}
\name{Eliya Nachmani$^{1,2}$,  Lior Wolf$^{1,2}$}
\address{$^{1}$Facebook AI Research, $^{2}$Tel-Aviv University}

\begin{document}
%
\maketitle
\begin{abstract}
We present a TTS neural network that is able to produce speech in multiple languages. The proposed network is able to transfer a voice, which was presented as a sample in a source language, into one of several target languages. Training is done without using matching or parallel data, i.e., without samples of the same speaker in multiple languages, making the method much more applicable. The conversion is based on learning a polyglot network that has multiple per-language sub-networks and adding loss terms that preserve the speaker's identity in multiple languages. We evaluate the proposed polyglot neural network for three languages with a total of more than 400 speakers and demonstrate convincing conversion capabilities.
\end{abstract}

\section{Introduction}
\label{sec:intro}

Neural text to speech (TTS) is an emerging technology that is becoming dominant over the alternative TTS technologies, in both quality and flexibility. On the quality front, multiple parallel lines of work, provide, in addition to increasing capabilities over time, an improved Mean Opinion Scores (MOS) from one work to the next: (i) From WaveNet~\cite{wavenet} to Parallel WaveNet~\cite{parallelwavenet}, and its alternative~\cite{ping2018clarinet}, (ii) from DeepVoice~\cite{deepvoice1} to DeepVoice2~\cite{deepvoice2} and DeepVoice3~\cite{deepvoice3}, (iii) from Tacotron~\cite{tacotron} to Tacotron2~\cite{tacotron2}, and (iv) from VoiceLoop~\cite{taigman2017voice} to its recent feed-forward-fitting successor~\cite{nachmani2018fitting}.

There is a parallel effort to improve the ease in which new voices are sampled. The first systems were able to reproduce a single voice, as captured from long high-quality recordings. Subsequent systems provided multi-speaker capabilities~\cite{taigman2017voice,baiduClone,NeosapienceClone,GoogleClone}, then multi-speakers sampled from challenging sources, such as audiobooks, and even multi-speakers sampled ``in the wild'' in noisy and uncontrolled recordings~\cite{nachmani2018fitting}. 

Our work continues this line of evolution and takes the voice sampling capabilities a step further. We would like to take a sample of a speaker talking in one language and have her voice-avatar speak, as a native speaker in another language. This direction is worth pursuing for several reasons. First, one of the most attractive applications of TTS is automatic translation of spoken words. This translation is most authentic, when maintaining the voice identity of the original speaker. Second, it is a challenging AI problem that requires the disentanglement of voice characteristics from spoken language characteristics. 

The proposed method is based on training a single TTS engine, with many components that are shared between the languages. In addition, there are two types of language-specific components. The first is a per-language encoder, that embeds the input sequences of phonemes in a language-independent vector space. The second component is a network that, given a sample of the speaker's voice, encodes it in a shared voice-embedding space. In other words, while there is one network per dataset that extracts the speaker's embedding from a speech sample, this embedding space is shared for all languages. This universality of the embedding is enforced by a novel loss term that preserves the speaker's identity under language conversion. The results show convincing conversions between English, Spanish, and German.

\section{Related Work}
\label{sec:format}




The recent neural TTS systems are sequence to sequence methods, where the underlying methods differ. Wavenet~\cite{wavenet} employs CNNs with dilated convolutions. Char2Wav~\cite{char2wav} employs RNNs, the original Tacotron method contains multiple RNNs, convolutions and a highway network~\cite{NIPS2015_5850}. The subsequent Tacotron2~\cite{tacotron2} method replaced the highway networks with RNNs and directly predicts the residuals. Deep Voice (DV)~\cite{deepvoice1} and DV2~\cite{deepvoice2} employ bidirectional RNNs, multilayer fully connected networks and residual connections. DV3~\cite{deepvoice3} switched the architecture to a gated convolutional sequence to sequence architecture~\cite{gehring2017convs2s} and also incorporated the key-value attention mechanism of~\cite{NIPS2017_7181}. The VoiceLoop~\cite{taigman2017voice} method is based on a specific type of RNN, in which a shifting buffer is used to maintain the context.

Out of this list, the multi-speaker systems are DV2, DV3, the modified Tacotron by the authors of DV2, and the two versions of VoiceLoop. 
Chat2Wav has a public multi-speaker implementation, with an added speaker embedding. 



The input text is represented by one of a few possible alternative types of features. Tacotron, Tacotron2 and DV3 employ the input letters. Phonemes are used by Char2Wav, VoiceLoop, and some of the DV3 experiments. Learned linguistic features, including duration and pitch of phonemes, are used by WaveNet, Parallel WaveNet, DV1, and DV2. Our work, which requires the modeling of multiple languages, relies on phonemes. Phoneme dictionaries are available for many languages. While we employ the same phonemes for English, Spanish and German, the method is general enough to support incompatible phoneme representations. 

The best audio quality is produced by methods that employ WaveNet decoders in order to generate wave forms, including DV1, DV2, DV3, Tacotron2. Char2Wav and VoiceLoop employ World Vocoder Features~\cite{morise2016world} in order to represent the output, which has to go through a separate decoding process. Our work also produces these vocoder features, which allows us to efficiently train on multiple corpora, even if these are insufficient for training a WaveNet decoder.





The literature has a few examples of polyglot TTS systems. These systems are mostly concatenative or HMM-based. The approach given in \cite{black2004multilingual} is a concatenative method that is based on unit selection. Polyglot synthesis is achieved by mapping the phonemes from the source language into the target language. An earlier system that uses matching samples of a single speaker~\cite{traber1999multilingual} uses a concatenative system that is applied to a recording of a single speaker speaking in four languages. Polyglot TTS based on HMM, has been proposed in various papers ~\cite{wu2008cross,peng2010cross,gibson2010unsupervised}.

Recently, neural multilingual TTS was proposed in \cite{ming2017light}. The method uses an LSTM architecture and is trained on a single speaker, who is recorded speaking two languages in the training data. Another related LSTM-based work~\cite{li2016multi} is trained on six different languages, with eight speakers in total. While the model is multi-language, it is not Polyglot -- every speaker speaks the language of the training set.

In~\cite{fan2016speaker}, a neural multilingual TTS is proposed. A deep neural network is trained on samples from a pair of languages -- English and Mandarin. Three speakers are recorded, each speaking both languages. In the last section of that paper, a neural polyglot TTS is discussed and demonstrated in a specific setting: the Mandarin samples of one of the speakers were removed from the training set, and a polyglot synthesis in Mandarin of this speaker was generated. Our work does not employ shared speakers among the corpora used for the different languages.

\section{Polyglot Neural Synthesis}
\label{sec:majhead}

The technique we employ in order to achieve a multi-language network, transcends any specific TTS architecture. We recommend the following modifications of the existing systems: (i) replacing the speaker embedding with a learned network, as was done in~\cite{nachmani2018fitting}, in order to verify via a loss that the output audio is compatible with the speaker's embedding, (ii) using multiple such speaker embedding networks, one per language, (iii) using multiple language (e.g., text or phoneme) embeddings, and (iv) using a loss term for speaker preserving. At test time, conversion is performed by a mixed input that consists of a sample of the speaker's voice speaking in the source language and a text in the target language.
\vspace*{-1cm}
\subsection{The Architecture}
\label{ssec:subhead}
\vspace*{-0.2cm}
The specific implementation of our method is based on the VoiceLoop architecture~\cite{taigman2017voice}, which is a neural multi-speaker Text To Speech (TTS) system. There is no specific reason to use this system, except that it was shown to outperform Char2Wav~\cite{char2wav}, which is the only other multi-speaker TTS system, whose open code is currently available.

The VoiceLoop is a biologically motivated architecture, which employs a ``phonological'' buffer $S_t$, which combines both auditory and lingual information. The model contains multiple networks ($N_u$, $N_a$, $N_o$), whose input includes this buffer.  $N_u$ creates the new vector $u_{t}$ that is pushed to the buffer at time $t$, $N_a$ is an attention network that produces weights over the input text, and $N_o$ generates the next audio frame, $o_t$. At every time step, $u_t$ is inserted into the buffer, and another vector is discarded in a FIFO manner.

The model also employs two LUTs, $LUT_p$ and $LUT_s$. The first LUT is the embedding of the input phonemes, one vector per each possible phoneme. $LUT_s$ stores a different embedding vector $z_s$ to each speaker encountered during training. 

\begin{figure}[t]
\centering
\includegraphics[width=0.475\textwidth]{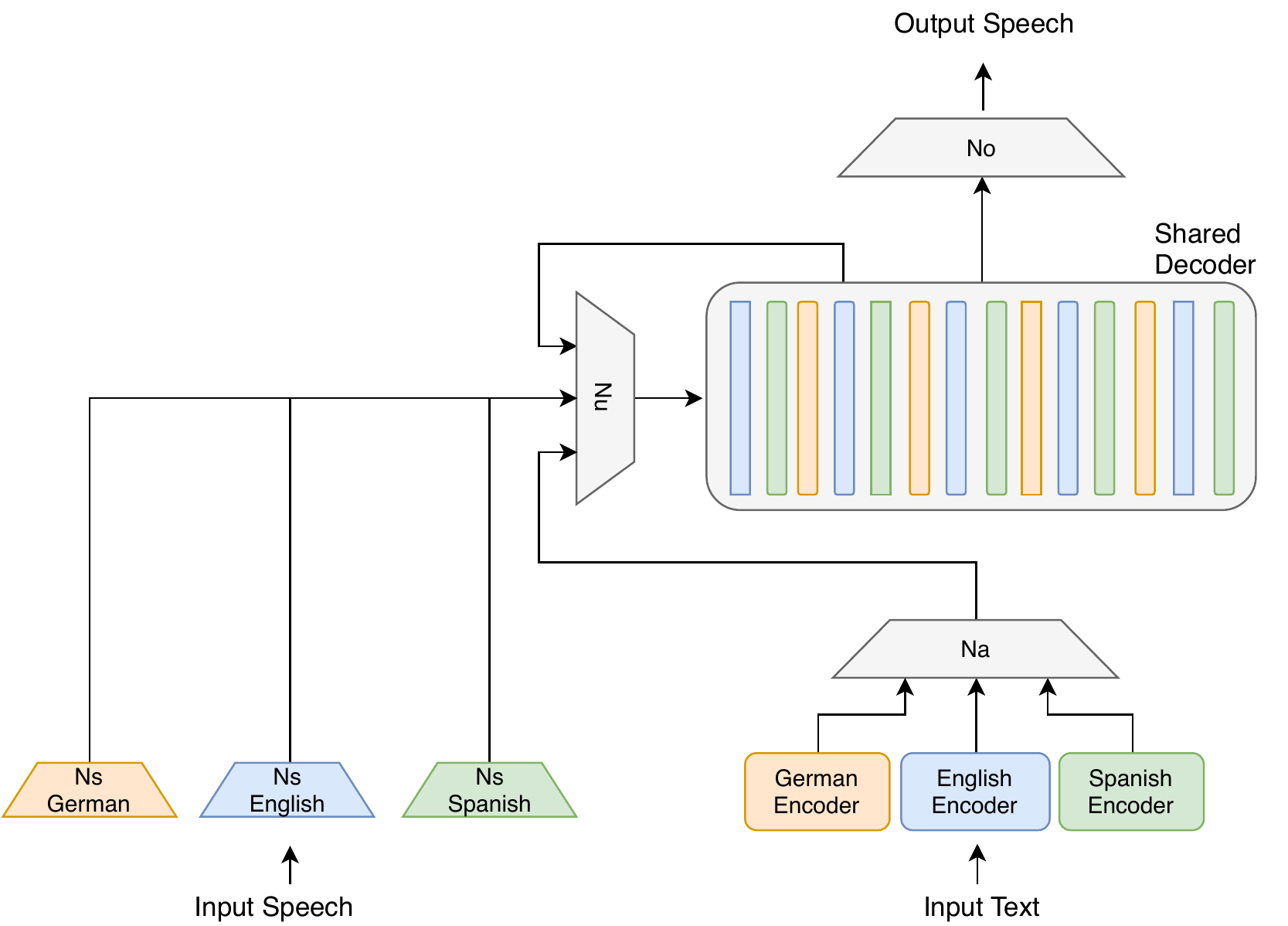}
\caption{The polyglot neural Text To Speech architecture.} 
	\label{fig:poly}
\end{figure}

We modify the architecture of~\cite{nachmani2018fitting}, by training a separate fitting network $N_s$ and a separate language embedding $LUT_p$ for each language. In order to achieve polyglot synthesis, we enforce a shared representation for all of the languages, and use only one copy of the attention network $N_a$, the buffer update $N_u$ and the output network $N_o$ (see Fig.~\ref{fig:poly}).

The architecture of the various networks follows~\cite{nachmani2018fitting}. The $N_s$ network has five conventional layers with ReLU activation. Each layer has $3 \times 3$ filters, with 32 channels. Average poling over time is performed after the conventional layers. The $N_s$ network ends with two fully connected layers with ReLU activation (each layer has size of $256$). The networks $N_o$, $N_u$ and $N_a$ have the same architecture: two fully connected layers with ReLU activations.
\vspace*{-1cm}
\subsection{The Loss Terms}
\label{sssec:subsubhead}

We propose a new loss function to achieve polyglot synthesis. Given voice sample $\mathbf{y^{a}}$ of the source language $\mathbf{a}$, and sequence of phoneme $\mathbf{s^{b}}$ from the target language $\mathbf{b}$. We would like to preserve the identity of the speaker in the voice sample created at the target language i.e., the embedding of the original speaker $\mathbf{z_s^{a}}$ should be the same as the embedding of the polyglot speech $\mathbf{z_s^{b}}$. An $L1$ loss term is used:
\begin{equation*}
L_{poly}=\sum_{\mathbf{a}}~\sum_{\mathbf{b}\neq \mathbf{a}}~\sum_{\mathbf{y^{a}},\mathbf{s^{b}}}\left \| \mathbf{z_s^{a}} - \mathbf{z_s^{b}} \right \|
\end{equation*}\vspace{-9pt}
\begin{equation*}
\mathbf{z_s^{a}}=N^a_s(\mathbf{y^{a}}),~\mathbf{z_s^{b}}=N^a_s\left ( \mathbf{{o}^{b}} \right ),~\mathbf{{o}^{b}}=G\left ( LUT_p^{b}(\mathbf{s^{b}}),\mathbf{z_s^{a}} \right )
\end{equation*}
where $G(LUT^{b}_p(\mathbf{s^b}),\mathbf{z_s})$ is the proposed polyglot network (a mapping between text in one language and voice sample in another to a new voice sample) $N^a_s$ is the sample embedding network of the source language $a$, and $LUT_p^{b}$ is the embedding of the text of the target language. Since we do not have matching parallel data, for this loss, unlike the training losses of VoiceLoop and VoiceLoop2, the network $G$ generates speech without teacher forcing. 
Note that for each speaker in the source language, we generate samples in the other languages, i.e. for each English speaker, we generate the converted voice in Spanish and German.

In addition to the loss above, we use the loss terms of VoiceLoop2~\cite{nachmani2018fitting}: (i) Reconstruction loss, given an audio sample $\mathbf{y}$, $L_{\text{MSE}} = \frac{1}{d_o}\sum_{\mathbf{y}}\sum_{t=1}^l \| y_t-o_t \|^2$, where the ground truth at time $t$ is $y_t$ and the network's output $G(LUT_p^a(s),N_s^a(y))$ at time $t$ is $o_t$, $s$ being the phoneme sequence of $y$ (both $y_t$ and $o_t$ are vectors in space of world vocoder features $\mathbb{R}^{d_o}$). (ii) A contrastive loss term that compares three samples from the same language $a$: $L_{\text{contrast}}  = \frac{1}{2}\sum_{\mathbf{y}^1,\mathbf{y}^2,\mathbf{y}^3}   (\|N_s(\mathbf{y}^1)-N_s(\mathbf{y}^2)\|^2 + \max(0,\Delta-\|N_s(\mathbf{y}^2)-N_s(\mathbf{y}^3)\|)^2)
$, where $\Delta$ is the margin, and $\mathbf{y}^1,\mathbf{y}^2,\mathbf{y}^3$ are voice samples, such that $\mathbf{y}^1,\mathbf{y}^2$ are from the same speaker, and $\mathbf{y}^3$ is a sample of another speaker of the same language. (iii) A cycle loss 
$L_{\text{cycle}} = \sum_{\mathbf{y}} \|N_s^a(\mathbf{y})-N_s^a(\mathbf{o})\|^2$, where $\mathbf{y}$ is the input audio sample of language $a$, and $\mathbf{o}$ is the output audio sample which is generated with $N^a_s(\mathbf{y})$ as embedding.

The overall loss is $L = L_{\text{MSE}} + \alpha L_{\text{contrast}}  + \beta L_{\text{cycle}} + \gamma L_{\text{poly}}$, where we set $\alpha=\beta=10$, and $\gamma=1000$

\subsection{Training and Inference Details}
\label{sec:print}
The $L_{poly}$ term is only effective when the network $G$ is able to generate speech in each one of the trained languages. Since it does not use teacher forcing, the signal it uses would diverge quickly for networks that are insufficiently trained. Therefore, the proposed network is trained with three phases, where the first two phases train the network to synthesize multilingual speech, and the third phase is for optimizing the speaker embedding space to enable convincing polyglot synthesis. 
\vspace*{-0.05cm}
Specifically, the first two phases follow the procedure, as described in~\cite{nachmani2018fitting}, except that the batches contain examples from multiple languages. The first phase uses white noise augmentation with standard deviation (SD) equal to 4.0 and sequences of ground truth vocoder features of length 100. The second (and the third) phases use SD of 2.0 and sequence length of 1000 vocoder features. The phases are run until convergence. The entire architecture is trained end-to-end on matching samples of inputs and outputs, where the input is the speaker embedding and the spoken text and the output is the recorded audio. The minimization of the MSE loss of the output frame, which is given as vocoder features, relies on the generated output to be perfectly aligned with the target, and is sensitive to drift. Teacher-forcing is, therefore, used, i.e., during training, the network receives the ground-truth target of the previous frame, in lieu of the predicted output from that frame.
\vspace*{-0.05cm}
The third phase of training adds the $L_{poly}$ term to the overall loss. Since the purpose of the third phase is to optimize the speaker embedding space to achieve polyglot synthesis, during this phase we only update the $N_s$ networks. 
At inference, polyglot synthesis is obtained by using $N^a_s$ of the source language $a$, and $LUT_p^{b}$ for the target language $b$. For example, we can synthesize an English speaker in the Spanish language, by using the speaker embedding from English $N^a_s$ and use as input a  sequence of Spanish phonemes, as encoded by the Spanish $LUT_p^{b}$. Note that the speaker does not need to be from the training set and samples from unseen speakers can be used for synthesis. Note also, that the transcript of the voice sample is not needed. However, the information on the language being spoken in that sample is required.
\vspace*{-0.05cm}
\section{Experiments}
\vspace*{-0.2cm}
\label{sec:page}


\begin{figure}[!tbp]
  \vspace{-.52cm}
  \begin{tabular}{cc}
    \includegraphics[width=.2\textwidth,height=0.2\textheight]{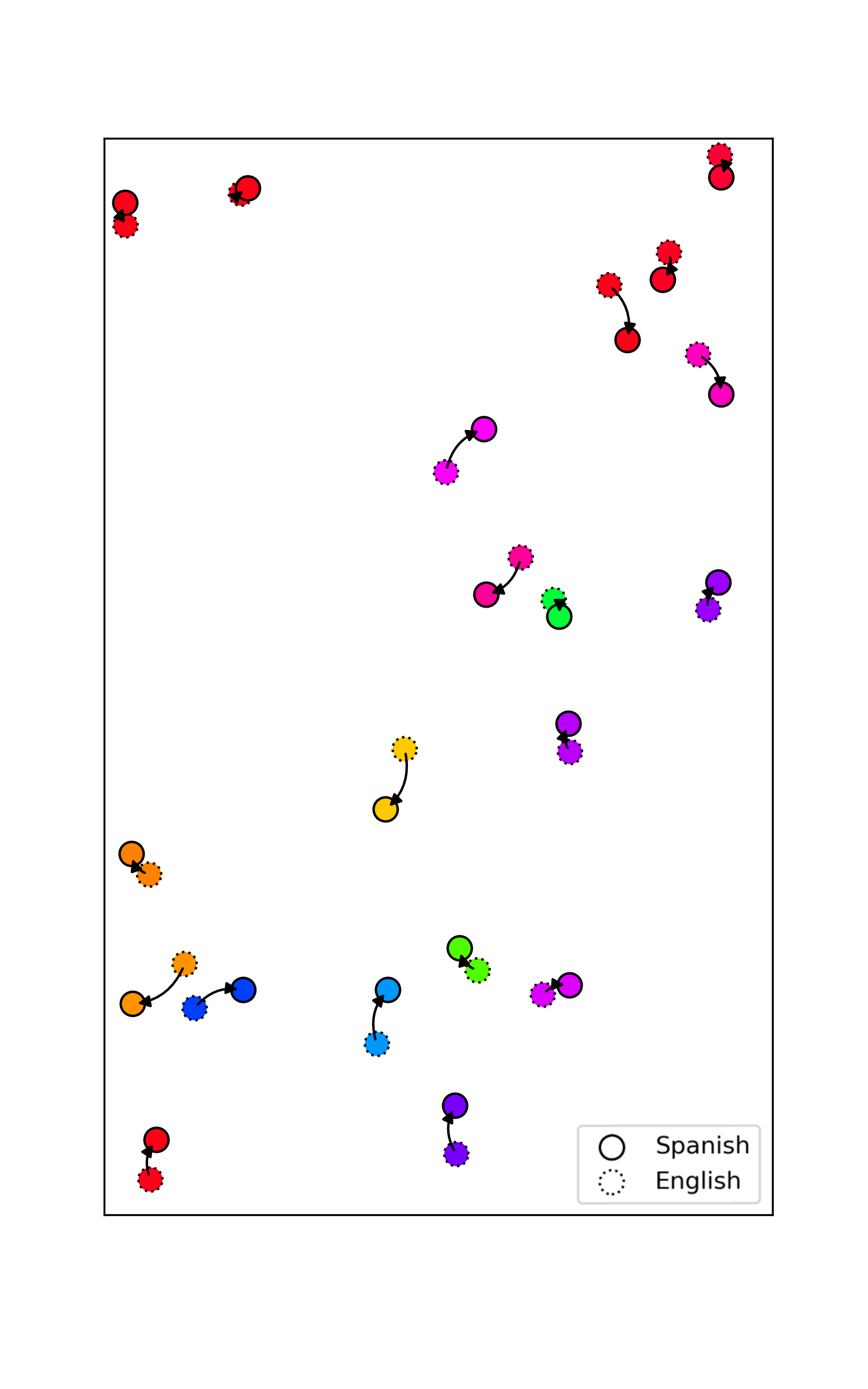} &
    \includegraphics[width=.2\textwidth,height=0.2\textheight]{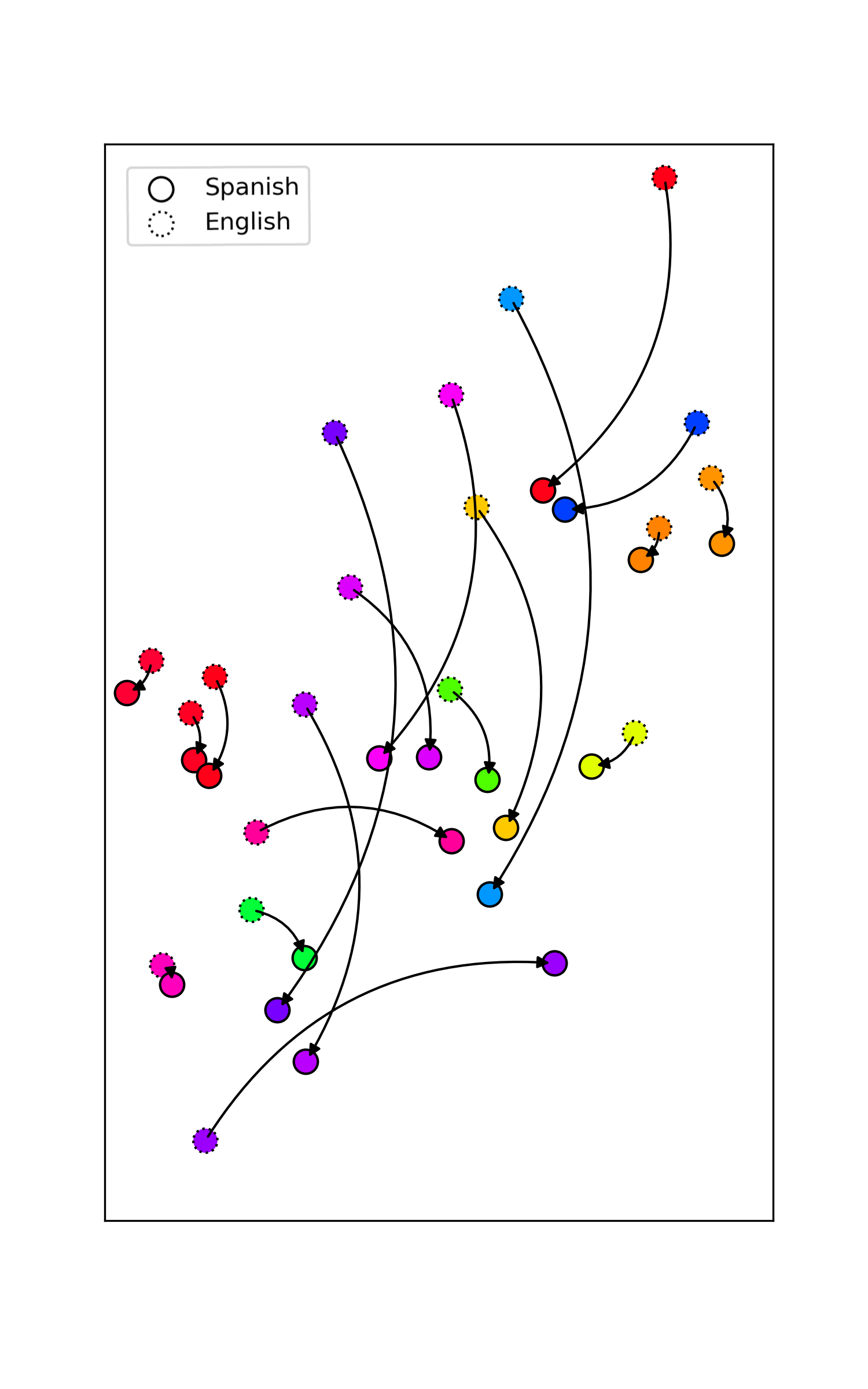}\vspace{-.5cm}
    \\
    
    (a) & (b)\\
    \end{tabular}
  \caption[Two numerical solutions]{TSNE visualization of the learned embedding for VCTK speakers in English and Spanish with (a) and without (b) the loss term $L_{poly}$. See text for details.}
  \label{fig:tsne} 
\end{figure}

We focus on three large multi-speaker datasets. For the English language, we use the VCTK dataset~\cite{vctk}, which contain 109 speakers. For the Spanish language, we use the DIMEx100 dataset \cite{pineda2010corpus}, which contain 100 speakers. For the German language, we use the PhonDat1 dataset \cite{PD1}, which contain 201 speakers. Audio samples can be found online~\url{https://ytaigman.github.io/polyglot/} and in the supplementary. 
It was shown in~\cite{nachmani2018fitting} that by embedding averaging the embeddings of many short samples, we improve the synthesis performance. We follow this technique, and for the speaker embedding during inference, we take a mean of $20$ training samples for each speaker, each of five seconds in length. 

In order to evaluate the quality of the generated audio, we employ Mean Opinion Score (MOS). All samples were presented at the 16kHz sample rate. The raters were told that they are presented with the results of the different algorithms. At least 10 raters participated in each such experiment. In addition, we measure the speaker similarity with two methods. The first method, evaluates similarity MOS. We provide pair of audio samples - a ground truth audio sample of the source language and a synthesized audio sample in the target language. The raters were asked to evaluate the similarity of speakers on a scale with $1-5$, were $1$ and $5$ scores corresponding to "different person", and "same person" respectively. The second method, uses a multiclass speaker identification deep network, employing the same identification neural network, as described  in~\cite{taigman2017voice}. 

In Tab.~\ref{tab:poly_nat}, we evaluate the naturalness of the generated speech sentences for the three languages. As can be seen, we get good results for MOS score quality (above $3.0$). However, it should be noted that the mean MOS score on the original samples from these datasets was about $4.46$. Following encoding and decoding with the world vocoder, this drops slightly to $4.19$.

In Tab.~\ref{tab:poly_sim}, we evaluate the similarity of the generated speech sentences for the three languages. As can be seen for the baseline, we get good self-similarity for the English language. Moreover, for English language with polyglot synthesis, we get good MOS similarity scores results, i.e. above $3.40$. Spanish and German obtains lower MOS similarity scores than English. We conjecture that this is a result of the fact that the VCTK dataset is larger then DIMEx100 and PhonDat1 (VCTK has $\sim40k$ voice samples, DIMEx100 has $\sim5.5k$ voice samples and PhonDat1 has $\sim15k$ voice samples). For the Spanish and German languages, we get good results for self-similarity (above $3.15$), and the Spanish to German (and vice versa) MOS scores polyglot synthesis results are close to the self-similarity MOS scores on these languages. 

In Tab.~\ref{tab:poly_sim_without}, we perform ablation analysis to show the contribution of the $L_{poly}$ term to the polyglot synthesis. Without this term, the polyglot behavior arises only from the shared embedding. As can be seen, we get acceptable MOS similarity scores results (above $3.0$). However, there is a noticeable gap in performance, in comparison to the results of the full system, as presented in Tab.~\ref{tab:poly_sim}. 

In Tab.~\ref{tab:poly_acc} and \ref{tab:poly_acc_no}, we evaluate the contribution of the $L_{poly}$ term on 
the automatic identification accuracy of the generated sentences. 
As expected, the best identification results are obtained for the self-language synthesis sentences (located along the diagonal). The polyglot synthesis accuracy (off diagonal) is good, in terms of both Top1 and Top5 accuracy (above $70\%$ and $91\%$, respectively). Where Top1 and Top5 refer to the accuracy of correctly classify the right speaker in the top one and the top five guesses, respectively. We can observe that adding $L_{poly}$ to the loss, improves the polyglot synthesis dramatically. For example, conversion from German to English improves in terms of Top1 accuracy from $82.71\%$ to $92.29\%$ when adding $L_{poly}$. Moreover, we can see that adding it also gets better or comparable results for same-language synthesis. 

In Fig.~\ref{fig:tsne}(a) (and \ref{fig:tsne}(b)), we provide t-Distributed Stochastic Neighbor Embedding (TSNE) visualization of the learned embedding for VCTK speakers in English and Spanish with (resp. without) the loss term $L_{poly}$. Each data point is the embedding, using $N^a_s$ of one voice sample, where $a$ is the sample's language. Data points with fragmented circles represent an original VCTK speaker in English, while data points with continuous circles represent the corresponding speaker in Spanish. As can be seen, adding $L_{poly}$ improves the invariance of the speakers' embedding to the language spoken.
\vspace*{-0.7cm}
\begin{table}[h]
\vspace{-.00cm}
\begin{small}
\centering
\caption{MOS naturalness scores (Mean $\pm$ SE)\label{tab:poly_nat}}
\begin{tabular}{lccc}
\toprule
\( Source \backslash Target \) & English & Spanish & German   \\
\midrule
English		& 3.05$\pm$1.30 & 3.01$\pm$1.34 & 3.30$\pm$1.15 \\
Spanish 	& 3.05$\pm$1.07 & 3.33$\pm$1.17 & 3.08$\pm$1.19 \\
German 	    & 3.35$\pm$0.99 & 3.12$\pm$1.08 & 3.02$\pm$1.00 \\
\bottomrule
\end{tabular}
\vspace{2pt}
\centering
\caption{MOS similarity scores (Mean $\pm$ SE)\label{tab:poly_sim}}
\begin{tabular}{lccc}
\toprule
\( Source \backslash Target \) & English & Spanish & German   \\
\midrule
English		& 3.51$\pm$1.21 &	3.41$\pm$1.37 &	3.19$\pm$1.36 \\
Spanish 	& 3.52$\pm$1.25 &	3.27$\pm$1.36 &	3.22$\pm$1.38 \\
German 	    & 3.40$\pm$1.25 &	3.15$\pm$1.35 &	3.16$\pm$1.39 \\
\bottomrule
\end{tabular}
\vspace{2pt}
\centering
\caption{No $L_{poly}$ - MOS similarity scores (Mean $\pm$ SE)\label{tab:poly_sim_without}}
\begin{tabular}{lccc}
\toprule
\( Source \backslash Target \) & English & Spanish & German   \\
\midrule
English		& 3.51$\pm$1.16 &	3.02$\pm$1.34 &	3.03$\pm$1.43 \\
Spanish 	& 3.00$\pm$1.37 &	3.01$\pm$1.43 &	3.13$\pm$1.33 \\ 
German 	    & 3.09$\pm$1.37 &	3.14$\pm$1.34 &	3.27$\pm$1.42 \\
\bottomrule
\end{tabular}
\vspace{2pt}
\centering
\caption{Identification accuracy top1[\%] (top5[\%])\label{tab:poly_acc}}
\begin{tabular}{lccc}
\toprule
\( Source\backslash Target \) & English & Spanish & German \\
\midrule
English		& 98.08 (99.66)	& 87.17 (94.57) & 84.32 (93.63)  \\
Spanish 	& 70.74 (91.62) & 89.88 (98.59)	& 70.02 (92.34)	 \\
German 	    & 92.29 (97.76)	& 90.67 (99.11)	& 94.40 (99.63)	 \\
\bottomrule
\end{tabular}
\vspace{2pt}
\caption{No $L_{poly}$ - Identification accuracy top1(top5)\label{tab:poly_acc_no}}
\begin{tabular}{lcccc}
\toprule
\( Source\backslash Target \) & English & Spanish & German \\
\midrule
English		& 97.96 (99.66) & 83.33 (91.67) & 80.44 (89.93) \\
Spanish 	& 66.47 (87.85)	& 89.91 (98.61)	& 67.21 (91.32) \\
German 	    & 82.71 (94.15)	& 88.44 (97.78)	& 94.28 (99.63) \\
\bottomrule
\end{tabular}
\end{small}
\end{table}


\vspace*{-1.0cm}
\section*{Acknowledgements}
\vspace*{-0.3cm}
 This work was carried out in partial fulfillment of the requirements for the Ph.D. degree of the first author.
 

\clearpage
\bibliographystyle{IEEEbib}
\bibliography{strings,refs_short}

\end{document}